\documentclass[11pt]{article}

\usepackage[utf8]{inputenc}
\usepackage[T1]{fontenc}
\usepackage{amsmath,amssymb,amsfonts}
\usepackage{graphicx}
\usepackage{booktabs}
\usepackage{hyperref}
\usepackage{natbib}
\usepackage{xcolor}
\usepackage{multirow}
\usepackage{caption}
\usepackage[margin=1in]{geometry}
\usepackage{algorithm}
\usepackage{algorithmic}
\usepackage{float}

\hypersetup{
    colorlinks=true,
    linkcolor=blue,
    citecolor=blue,
    urlcolor=blue
}

\title{PolyGLU: State-Conditional Activation Routing \\ in Transformer Feed-Forward Networks}

\author{
  Daniel Nobrega Medeiros\thanks{Independent researcher. Correspondence: \url{https://github.com/danielxmed/PolyGLU}} \\
  \\
  \small Code: \url{https://github.com/danielxmed/PolyGLU} \\
  \small Base model: \url{https://huggingface.co/tylerxdurden/PolyChromaticLM-1.0-base-0.6B} \\
  \small Instruct model: \url{https://huggingface.co/tylerxdurden/PolyChromaticLM-1.0-instruct-0.6B}
}

\date{March 2026}

\begin{document}
\maketitle

\begin{abstract}
Biological neural systems employ diverse neurotransmitters---glutamate, GABA, dopamine, acetylcholine---to implement distinct signal-processing modalities within shared neural circuits. In contrast, modern transformers apply a single fixed activation function across all feed-forward neurons. We introduce \textbf{PolyGLU} (Polychromatic Gated Linear Unit), a drop-in replacement for SwiGLU that enables each FFN neuron to dynamically route among $K{=}4$ activation functions via a differentiable mechanism combining learned static preferences with input-conditioned gating, trained end-to-end with Gumbel-Softmax. We train PolychromaticLM, a 597M-parameter transformer, on ${\sim}10$B tokens using a single NVIDIA A100 GPU. Our key finding is \emph{emergent routing behavior}: without any explicit sparsity loss or entropy regularization, the routing mechanism converges to near-deterministic activation selections (mean dynamic entropy = 0.030\% of maximum), with a striking depth-dependent specialization pattern---early layers prefer GELU while deep layers strongly favor Tanh. Three layers maintain elevated routing entropy, suggesting computational flexibility points. The routing architecture adds only 0.23\% parameter overhead (${\sim}1.4$M parameters) and proves fully robust to supervised fine-tuning: routing entropy remains constant at $\ln(4)$ throughout 13,067 SFT steps. On standard benchmarks, PolychromaticLM achieves 62--89\% of Qwen3-0.6B-Base performance despite training on 3{,}600$\times$ fewer tokens. All code, weights, and training infrastructure are released under Apache~2.0.
\end{abstract}

\section{Introduction}

Biological neural systems do not rely on a single signaling mechanism. Instead, they employ a diverse repertoire of neurotransmitters---each conferring distinct computational properties on the circuits they modulate \citep{kandel2013principles}. Glutamate provides fast excitatory transmission, GABA mediates inhibition, dopamine gates reward-modulated learning, and acetylcholine regulates attentional allocation. This diversity is not incidental; it is a fundamental architectural feature that enables the brain to implement qualitatively different computations within a shared neural substrate.

Modern transformer architectures \citep{vaswani2017attention}, by contrast, apply a single fixed activation function uniformly across all feed-forward neurons. The evolution from ReLU \citep{nair2010relu} to GELU \citep{hendrycks2016gelu} to SwiGLU \citep{shazeer2020glu} has improved performance, but the fundamental assumption remains: one activation function is optimal for all neurons at all depths for all inputs.

We challenge this assumption with \textbf{PolyGLU} (Polychromatic Gated Linear Unit), a drop-in SwiGLU replacement that allows each FFN neuron to dynamically select among $K{=}4$ candidate activation functions---ReLU, Tanh, SiLU, and GELU---through a differentiable routing mechanism. Each neuron maintains a learned static preference over the activation palette, modulated by a lightweight gating network conditioned on the current hidden state. Routing decisions are made differentiable via Gumbel-Softmax \citep{jang2017categorical, maddison2017concrete} with temperature annealing.

Our primary contribution is not the mechanism itself, but the \emph{emergent behavior} it produces. When we train PolychromaticLM---a 597M-parameter transformer with PolyGLU---on ${\sim}10$B tokens, we observe:

\begin{enumerate}
    \item \textbf{Spontaneous routing convergence.} Without any explicit sparsity loss, entropy penalty, or load-balancing regularizer, the routing mechanism converges to near-deterministic selections (mean dynamic entropy = 0.030\% of maximum). The model \emph{discovers} that sparse, committed activation routing is preferable to soft mixing.

    \item \textbf{Depth-dependent specialization.} A clear activation gradient emerges across the 28 transformer layers: early layers predominantly select GELU (probabilistic gating), while deep layers strongly favor Tanh (bounded compression). This learned specialization suggests that different network depths require different nonlinear transformations---a finding that could inform future architecture design.

    \item \textbf{Fine-tuning robustness.} During supervised fine-tuning on mathematical reasoning data (13,067 steps), routing entropy remains \emph{exactly} constant at $\ln(4)$, indicating that the routing architecture cleanly separates ``how to compute'' from ``what to compute.'' PolyGLU models can be safely fine-tuned without risk of routing collapse or activation degeneration.
\end{enumerate}

All of this is achieved with only 0.23\% parameter overhead (${\sim}$1.4M routing parameters out of 597M total), demonstrating that activation routing can be made practical at negligible cost.

This work was conducted as independent research on a single NVIDIA A100 80GB GPU rented from RunPod community cloud at ${\sim}$\$1.64/hr, with a total project budget of approximately \$346. We view this resource constraint as a feature rather than a limitation: it demonstrates that novel architectural research can be conducted outside of large industrial labs, and all results are straightforwardly reproducible on commodity hardware. The model, weights, and all training code are fully open-source under Apache~2.0.\footnote{Code: \url{https://github.com/danielxmed/PolyGLU}. Base model: \url{https://huggingface.co/tylerxdurden/PolyChromaticLM-1.0-base-0.6B}. Instruct model: \url{https://huggingface.co/tylerxdurden/PolyChromaticLM-1.0-instruct-0.6B}.}

\section{Related Work}

\paragraph{Activation Functions.}
The history of activation functions in neural networks traces a path from hard thresholds to smooth, gated nonlinearities. ReLU \citep{nair2010relu} replaced saturating activations with a simple thresholding operation. GELU \citep{hendrycks2016gelu} introduced probabilistic gating. The Gated Linear Unit (GLU) family \citep{dauphin2017language} showed that element-wise gating in feed-forward layers improves language modeling, leading to SwiGLU \citep{shazeer2020glu}---now the dominant FFN activation in modern transformers including LLaMA and Qwen \citep{yang2025qwen3}. PolyGLU generalizes this line by replacing the fixed activation in SwiGLU with a learned, input-conditioned selection among multiple candidates.

\paragraph{Mixture of Experts.}
Sparsely-activated models \citep{shazeer2017outrageously, fedus2022switch} route entire tokens to different expert sub-networks, achieving parameter efficiency through conditional computation. PolyGLU operates at a finer granularity: rather than routing tokens to experts, it routes individual neurons to activation functions. This is more analogous to neurotransmitter selection in biological neurons than to expert assignment. Notably, Mixture-of-Experts models require explicit load-balancing losses to prevent routing collapse, whereas PolyGLU achieves near-deterministic routing \emph{without} any auxiliary loss.

\paragraph{Gumbel-Softmax.}
The Gumbel-Softmax estimator \citep{jang2017categorical, maddison2017concrete} enables gradient-based optimization through discrete categorical choices by providing a continuous relaxation. It has been applied to learned quantization, discrete latent variable models, and architecture search. We use it as the differentiable routing mechanism for activation selection, with temperature annealing from $\tau{=}1.0$ (exploration) to $\tau{=}0.1$ (commitment).

\paragraph{Adaptive Computation.}
Prior work on adaptive computation has explored input-dependent depth \citep{graves2016adaptive}, width, or routing. PolyGLU is conceptually closest to work on adaptive activation functions, but differs in that the routing is \emph{per-neuron} and \emph{per-layer} rather than global, and the routing signal combines both static learned preferences and dynamic input conditioning.

\section{Method}

\subsection{Background: SwiGLU}

The SwiGLU feed-forward block \citep{shazeer2020glu} computes:
\begin{equation}
    \text{SwiGLU}(\mathbf{x}) = \left[\text{SiLU}(\mathbf{x} W_{\text{gate}})\right] \odot (\mathbf{x} W_{\text{up}})
\end{equation}
followed by a down-projection $W_{\text{down}}$. The SiLU (Swish) activation is applied identically to every neuron in every layer.

\subsection{PolyGLU Formulation}

PolyGLU generalizes SwiGLU by replacing the fixed SiLU with a learned mixture of $K$ activation functions:
\begin{equation}
    \text{PolyGLU}(\mathbf{x}) = \left[\sum_{k=1}^{K} g_k \cdot \sigma_k(\mathbf{x} W_{\text{gate}})\right] \odot (\mathbf{x} W_{\text{up}})
    \label{eq:polyglu}
\end{equation}
where $\sigma_k$ are the candidate activation functions and $g_k$ are per-neuron routing weights computed via a two-component routing mechanism.

We use $K{=}4$ activation functions, chosen to span qualitatively different nonlinear behaviors:

\begin{table}[h]
\centering
\small
\begin{tabular}{@{}clll@{}}
\toprule
$k$ & Function & Property & Biological Analogy \\
\midrule
0 & ReLU & Hard threshold & Glutamate (excitatory) \\
1 & Tanh & Symmetric compression & GABA (inhibitory) \\
2 & SiLU & Self-gated & Dopamine (modulatory) \\
3 & GELU & Probabilistic gate & Acetylcholine (attentional) \\
\bottomrule
\end{tabular}
\caption{PolyGLU activation palette. The biological analogies are inspirational, not claims of functional equivalence.}
\label{tab:activations}
\end{table}

\subsection{Routing Mechanism}

The routing weights $g_k$ are computed from two components---a static preference and a dynamic, input-conditioned modulation:

\paragraph{Static preferences.} Each neuron $j$ (of $d_{\text{ff}}$ total) maintains a learnable preference vector $\boldsymbol{\alpha}_j \in \mathbb{R}^K$, initialized to zero (uniform prior). These encode baseline activation affinities analogous to a neuron's intrinsic neurotransmitter identity.

\paragraph{Dynamic gating.} A lightweight MLP processes the mean-pooled hidden state $\bar{\mathbf{h}} = \text{mean}(\mathbf{x}, \text{dim}{=}\text{seq})$ to produce context-dependent routing modulation:
\begin{equation}
    f(\bar{\mathbf{h}}) = W_2 \cdot \text{ReLU}(W_1 \bar{\mathbf{h}} + b_1) + b_2
\end{equation}
where $W_1 \in \mathbb{R}^{32 \times d_{\text{model}}}$ and $W_2 \in \mathbb{R}^{K \times 32}$. Per-activation scaling factors $\boldsymbol{\beta} \in \mathbb{R}^K$ (initialized to 1.0) modulate the dynamic signal.

\paragraph{Combined routing.} The full routing logits combine both components:
\begin{equation}
    \ell_k = \alpha_k + \beta_k \cdot f(\bar{\mathbf{h}})_k
\end{equation}
\begin{equation}
    g_k = \text{GumbelSoftmax}(\boldsymbol{\ell}, \tau)_k
    \label{eq:routing}
\end{equation}

The Gumbel-Softmax temperature $\tau$ is annealed linearly during training:
\begin{equation}
    \tau(t) = \max\!\left(0.1,\; 1.0 - 0.9 \cdot \frac{t}{t_{\text{total}}}\right)
    \label{eq:tau}
\end{equation}

At $\tau{=}1.0$ (training start), routing is nearly uniform, encouraging exploration. At $\tau{=}0.1$ (training end), routing is near-deterministic, encouraging commitment.

\subsection{Integration into the Transformer Block}

Each transformer block follows a pre-norm residual structure:
\begin{align}
    \mathbf{x} &\leftarrow \mathbf{x} + \text{GQA}(\text{RMSNorm}(\mathbf{x})) \\
    \mathbf{x} &\leftarrow \mathbf{x} + \text{PolyGLU}(\text{RMSNorm}(\mathbf{x}))
\end{align}
PolyGLU is a drop-in replacement: $W_{\text{gate}}$, $W_{\text{up}}$, and $W_{\text{down}}$ retain their standard dimensions. The routing mechanism adds only the $\boldsymbol{\alpha}$, $\boldsymbol{\beta}$, and gate network parameters---approximately 1.4M additional parameters (0.23\% overhead) for a 597M-parameter model.

\subsection{Parameter Overhead Analysis}

Per PolyGLU layer, the routing overhead is:
\begin{itemize}
    \item $\boldsymbol{\alpha} \in \mathbb{R}^{d_{\text{ff}} \times K}$: $4{,}096 \times 4 = 16{,}384$ parameters
    \item $\boldsymbol{\beta} \in \mathbb{R}^K$: $4$ parameters
    \item Gate network: $\text{Linear}(1024 \to 32) + \text{Linear}(32 \to 4)$: $32{,}768 + 32 + 128 + 4 = 32{,}932$ parameters
    \item \textbf{Total per layer}: ${\sim}49{,}320$ parameters
    \item \textbf{Total (28 layers)}: ${\sim}1.4$M parameters (0.23\% of 597M)
\end{itemize}

\section{Experimental Setup}

\subsection{Model Architecture}

PolychromaticLM is a decoder-only transformer with the specifications in Table~\ref{tab:architecture}.

\begin{table}[h]
\centering
\small
\begin{tabular}{@{}lr@{}}
\toprule
\textbf{Parameter} & \textbf{Value} \\
\midrule
Total parameters & 597,153,888 \\
\quad of which routing & ${\sim}$1.4M (0.23\%) \\
Hidden dimension ($d_{\text{model}}$) & 1,024 \\
FFN intermediate ($d_{\text{ff}}$) & 4,096 \\
Layers & 28 \\
Query / KV heads & 16 / 8 (GQA) \\
Head dimension & 64 \\
Context length & 4,096 tokens \\
Vocabulary & 151,669 (Qwen3 tokenizer) \\
Position encoding & RoPE ($\theta{=}10{,}000$) \\
Normalization & RMSNorm (pre-norm) + QK-Norm \\
FFN activation & PolyGLU ($K{=}4$) \\
Weight tying & Embedding $\leftrightarrow$ output head \\
\bottomrule
\end{tabular}
\caption{PolychromaticLM architecture. The model uses Grouped Query Attention \citep{ainslie2023gqa} with RoPE \citep{su2024roformer} and RMSNorm \citep{zhang2019root}.}
\label{tab:architecture}
\end{table}

Residual connections to $W_o$ (attention output) and $W_{\text{down}}$ (FFN output) are scaled by $1/\sqrt{2 \cdot n_{\text{layers}}} \approx 0.134$. Weight initialization follows $\mathcal{N}(0, 0.02)$ for linear layers and embeddings, with $\boldsymbol{\alpha}$ initialized to zero (uniform prior) and $\boldsymbol{\beta}$ initialized to ones.

\subsection{Pre-Training}

\paragraph{Data.} The model was trained on ${\sim}$10.24B tokens from three domains:

\begin{table}[h]
\centering
\small
\begin{tabular}{@{}llrr@{}}
\toprule
\textbf{Domain} & \textbf{Dataset} & \textbf{Share} & \textbf{Tokens} \\
\midrule
Math & \texttt{nvidia/Nemotron-CC-Math-v1} (4+) & 70\% & 7.0B \\
STEM & \texttt{openbmb/Ultra-FineWeb} & 25\% & 2.5B \\
Code & \texttt{lumees/github-code-2025} (Python) & 5\% & 0.5B \\
\midrule
\textbf{Total} & & & \textbf{10.0B} \\
\bottomrule
\end{tabular}
\caption{Pre-training data mix. A two-phase schedule anneals the mix from 70/25/5 to 85/10/5 (math/STEM/code) over the final 20\% of training.}
\label{tab:data}
\end{table}

All data was tokenized using the Qwen3 tokenizer (\texttt{Qwen/Qwen3-0.6B-Base}, vocabulary 151,669) into binary uint32 chunks of ${\sim}$100M tokens. Documents within chunks are separated by EOS tokens (ID 151643), and document masking is enforced via Flash Attention 2's \citep{dao2022flashattention} variable-length mode (\texttt{flash\_attn\_varlen\_func}), preventing cross-document attention.

\paragraph{Optimization.} We use AdamW \citep{loshchilov2019decoupled} with $\beta_1{=}0.9$, $\beta_2{=}0.95$, $\epsilon{=}10^{-8}$, weight decay 0.1 (applied to 2D+ weight matrices only; biases, norms, $\boldsymbol{\alpha}$, and $\boldsymbol{\beta}$ are exempt). The learning rate follows cosine decay with a 2,000-step linear warmup to a peak of $10^{-4}$. Gradient clipping is set to 1.0 (max norm).

\paragraph{Batching.} Micro-batch size of 16 sequences $\times$ 4,096 tokens $\times$ 8 gradient accumulation steps = 524,288 tokens per effective batch. Training runs for 19,531 steps (${\sim}$10.24B tokens total).

\paragraph{Infrastructure.} All training was conducted on a single NVIDIA A100 80GB GPU with DeepSpeed \citep{rasley2020deepspeed} ZeRO Stage~0, BFloat16 precision, gradient checkpointing, and chunked cross-entropy loss (avoiding materialization of the full $[B {\times} T {\times} 151{,}669]$ logit tensor). Mean throughput: ${\sim}$11,800 tokens/sec. Total wall time: ${\sim}$12.5 days.

\subsection{Supervised Fine-Tuning}

The pre-trained model was fine-tuned on \texttt{nvidia/Nemotron-Math-v2} (${\sim}$347K math problems in ChatML format) for 1 epoch (13,067 steps). Loss was computed on assistant tokens only. Learning rate: $2 \times 10^{-5}$ with cosine decay and 100-step warmup. The Gumbel-Softmax temperature was frozen at $\tau{=}0.1$. Training completed in ${\sim}$18 hours (${\sim}$\$29.50 compute cost).

\subsection{Evaluation Protocol}

We use the EleutherAI lm-evaluation-harness \citep{eval-harness} v0.4.11 with default settings. Our benchmark suite comprises 10 loglikelihood tasks (Table~\ref{tab:benchmarks}), evaluated in zero-shot unless noted. We additionally compute domain perplexity on held-out data from each training domain (244 non-overlapping sequences of 4,096 tokens per domain).

Our primary reference model is \textbf{Qwen3-0.6B-Base} \citep{yang2025qwen3}: same tokenizer, comparable parameter count (${\sim}$600M), but trained on ${\sim}$36T tokens---approximately 3,600$\times$ our budget.

\section{Results}

\subsection{Pre-Training Convergence}

\begin{figure}[h]
    \centering
    \includegraphics[width=0.85\linewidth]{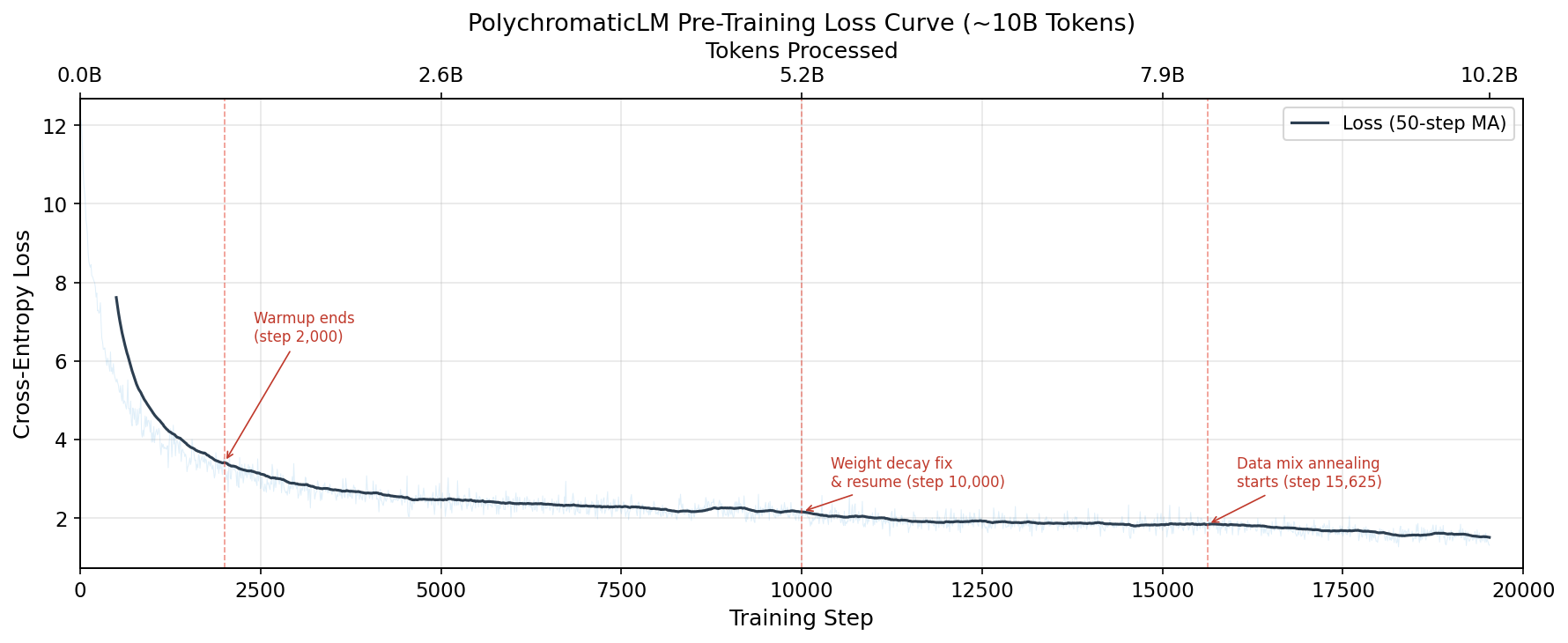}
    \caption{Pre-training loss curve. Loss decreases from 12.13 to 1.31 over ${\sim}$10.24B tokens (19,531 steps). A mid-training intervention at step 10,000 (Section~\ref{sec:weight_decay}) introduced no visible discontinuity.}
    \label{fig:loss_curve}
\end{figure}

Training loss decreased from 12.13 to 1.31 over 19,531 steps (Figure~\ref{fig:loss_curve}), representing an 89\% reduction. The loss curve exhibits three phases: rapid descent during warmup (steps 0--2,000), steady optimization (steps 2,000--15,625), and fine convergence during data mix annealing (steps 15,625--19,531). Throughput remained stable at ${\sim}$11,800 tokens/sec throughout.

\begin{figure}[h]
    \centering
    \includegraphics[width=0.85\linewidth]{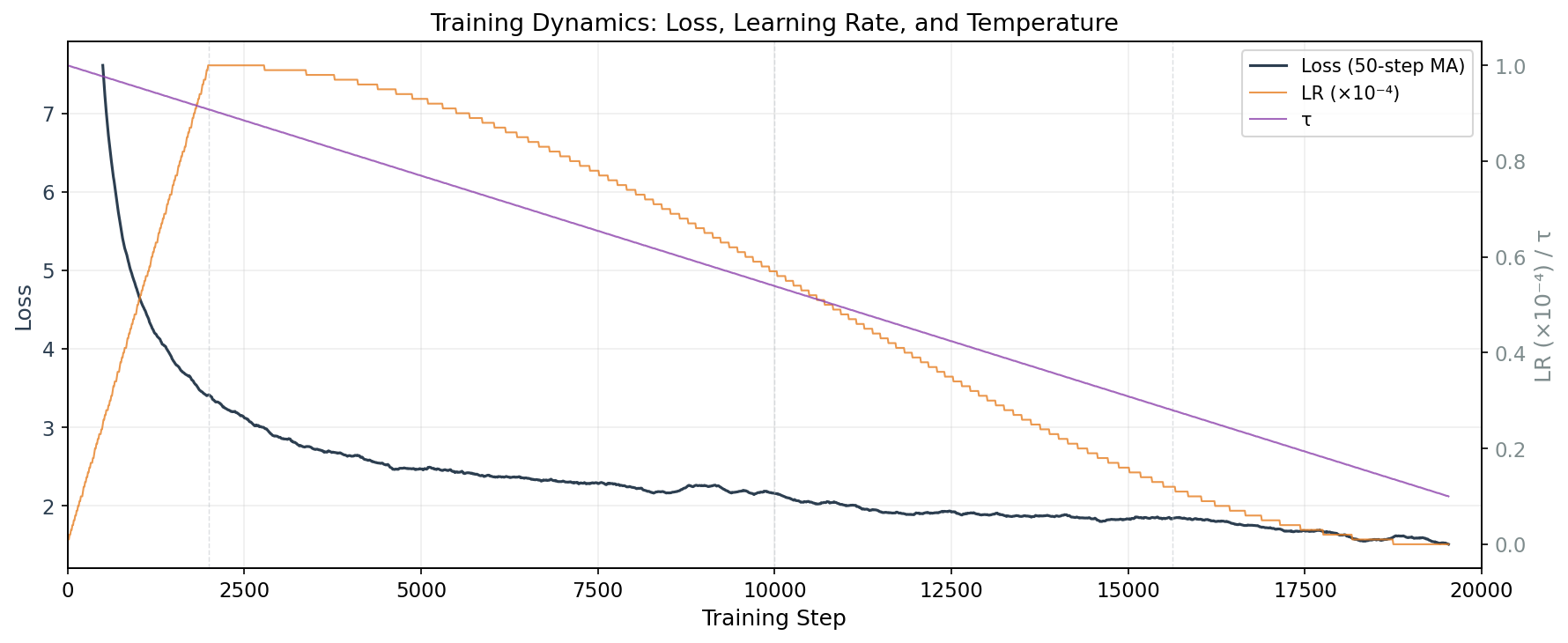}
    \caption{Combined training dynamics showing loss, learning rate, Gumbel-Softmax temperature ($\tau$), and throughput over the full training run.}
    \label{fig:combined_dynamics}
\end{figure}

\subsection{Emergent Routing Behavior}
\label{sec:routing}

The central finding of this work is that PolyGLU's routing mechanism converges to near-deterministic activation selections \emph{without any explicit regularization}. No sparsity loss, entropy penalty, or load-balancing auxiliary objective was applied during training---the routing signal emerges purely from the language modeling objective.

\subsubsection{Near-Deterministic Convergence}

\begin{figure}[h]
    \centering
    \includegraphics[width=0.85\linewidth]{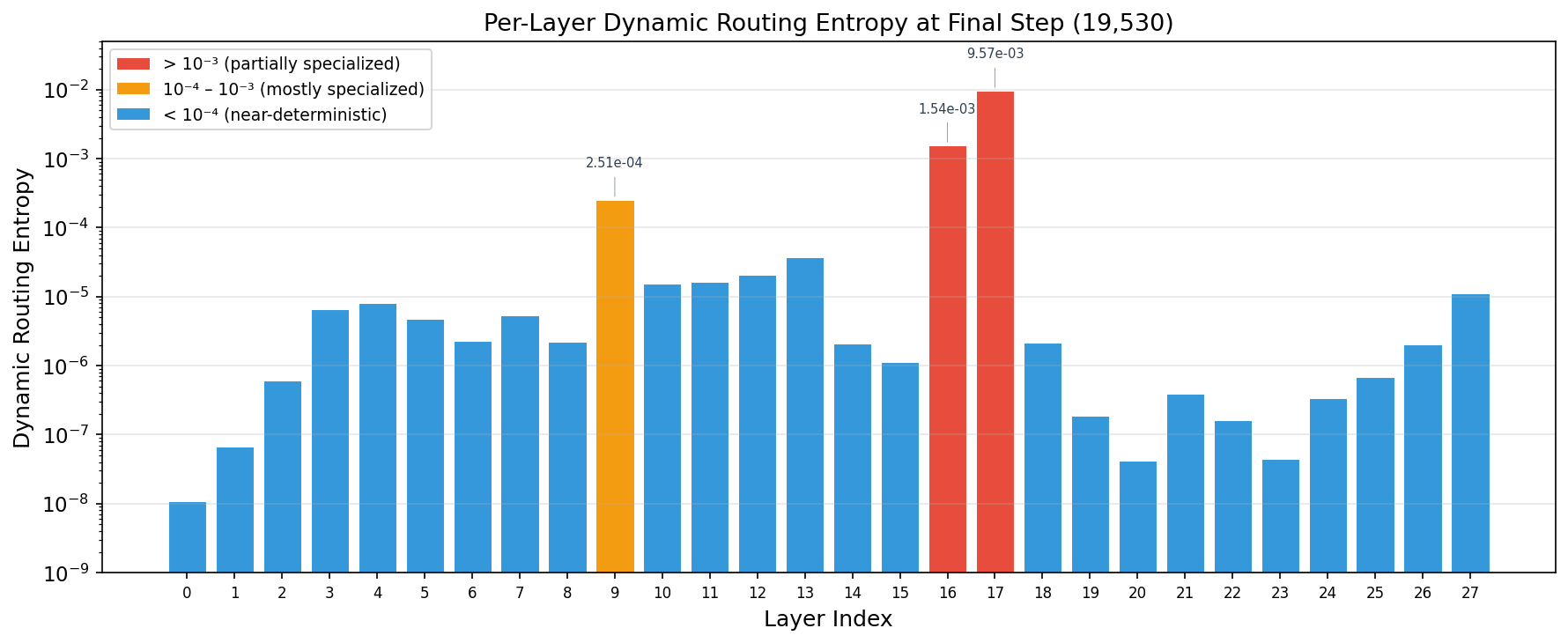}
    \caption{Per-layer dynamic routing entropy at convergence (step 19,531). Most layers achieve entropy $< 10^{-4}$, with three notable exceptions: layers 9, 16, and 17 maintain elevated entropy, suggesting computational flexibility points.}
    \label{fig:entropy_final}
\end{figure}

We measure routing behavior via \textit{dynamic routing entropy}: the entropy of the softmax distribution over routing logits $(\boldsymbol{\alpha} + \boldsymbol{\beta} \cdot f(\bar{\mathbf{h}}))$ computed on held-out data. At convergence (step 19,531, $\tau{=}0.1$), the mean dynamic routing entropy across all 28 layers is $4.1 \times 10^{-4}$, representing only \textbf{0.030\%} of the theoretical maximum $\ln(4) \approx 1.386$ (Figure~\ref{fig:entropy_final}).

The entropy is not uniform across layers. Most layers (0--8, 10--15, 18--27) achieve entropy below $10^{-4}$, effectively making one-hot activation selections. Three layers stand out:
\begin{itemize}
    \item \textbf{Layer 9}: entropy $2.5 \times 10^{-4}$---modestly elevated
    \item \textbf{Layer 16}: entropy $1.5 \times 10^{-3}$---partially specialized
    \item \textbf{Layer 17}: entropy $9.6 \times 10^{-3}$---the highest in the network
\end{itemize}

Layer 17 is particularly interesting: its entropy \emph{increased} during the final training phase (from $6.9 \times 10^{-4}$ at step 10,030 to $9.6 \times 10^{-3}$ at step 19,531), counter to the global trend. This suggests layer 17 actively benefits from maintaining activation diversity.

\begin{figure}[h]
    \centering
    \includegraphics[width=0.85\linewidth]{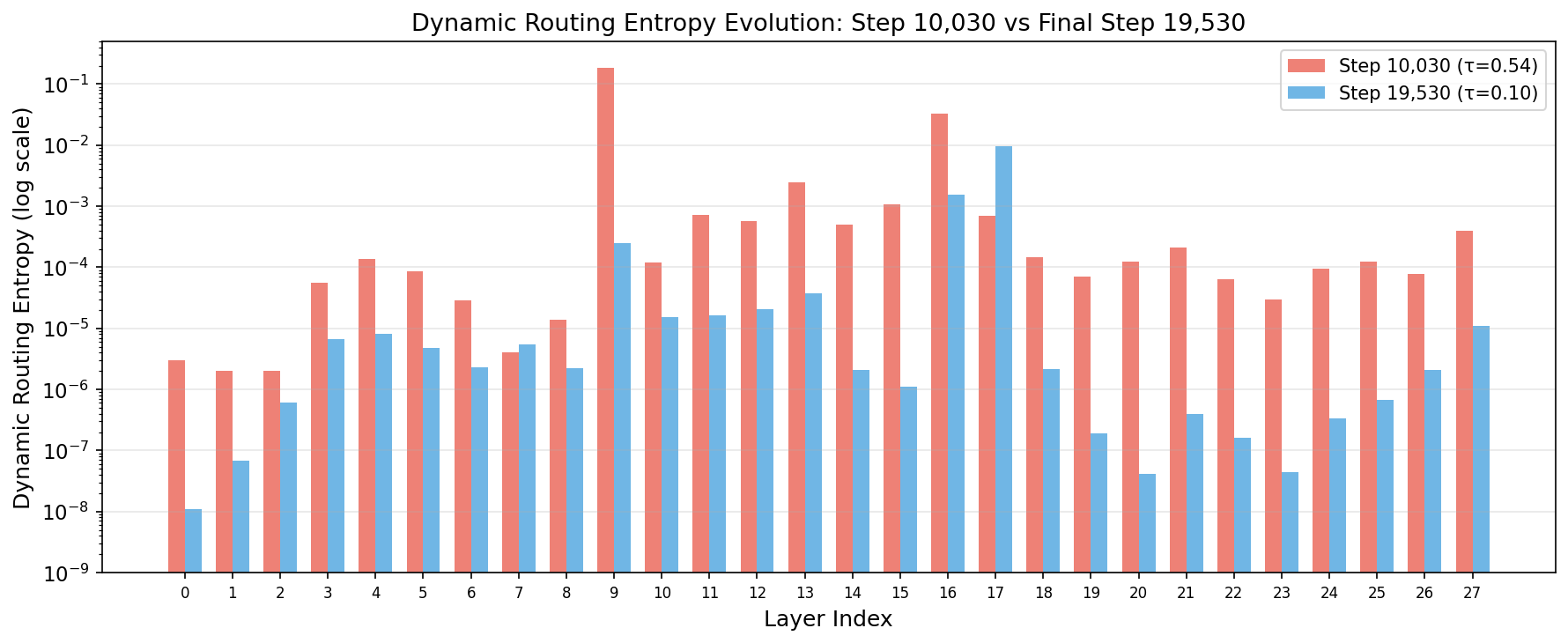}
    \caption{Evolution of dynamic routing entropy during training. Most layers converge to near-zero entropy, while layers 9, 16, and 17 maintain elevated values. Layer 17 notably \emph{increases} its entropy in the final phase.}
    \label{fig:entropy_evolution}
\end{figure}

\subsubsection{Layer-Wise Activation Specialization}

\begin{figure}[t]
    \centering
    \includegraphics[width=\linewidth]{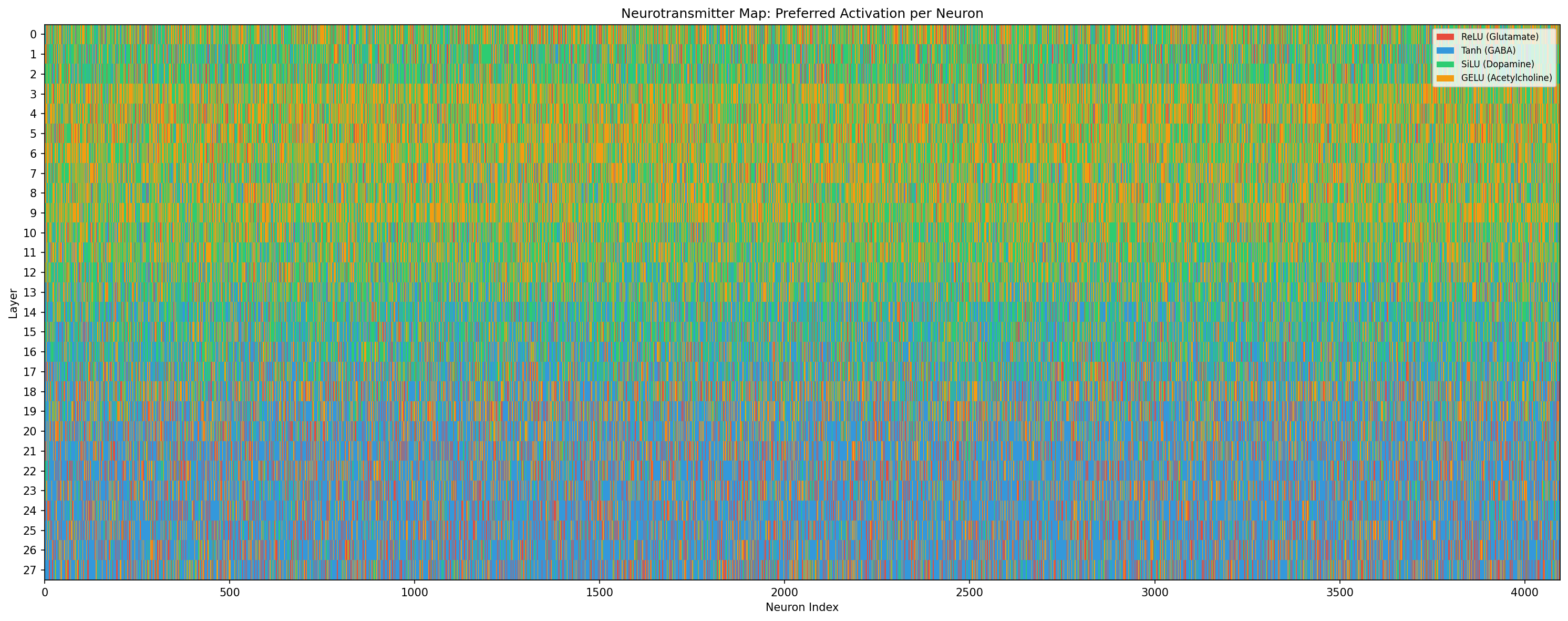}
    \caption{Neurotransmitter heatmap showing the preferred activation function per neuron (columns) across layers (rows). A clear depth-dependent specialization emerges: early layers favor GELU (blue), while deep layers strongly prefer Tanh (orange).}
    \label{fig:heatmap}
\end{figure}

The neurotransmitter heatmap (Figure~\ref{fig:heatmap}) visualizes the preferred activation function ($\arg\max_k \alpha_k$) for each of the 4,096 neurons across 28 layers, revealing a striking depth-dependent gradient.

\begin{figure}[h]
    \centering
    \includegraphics[width=0.85\linewidth]{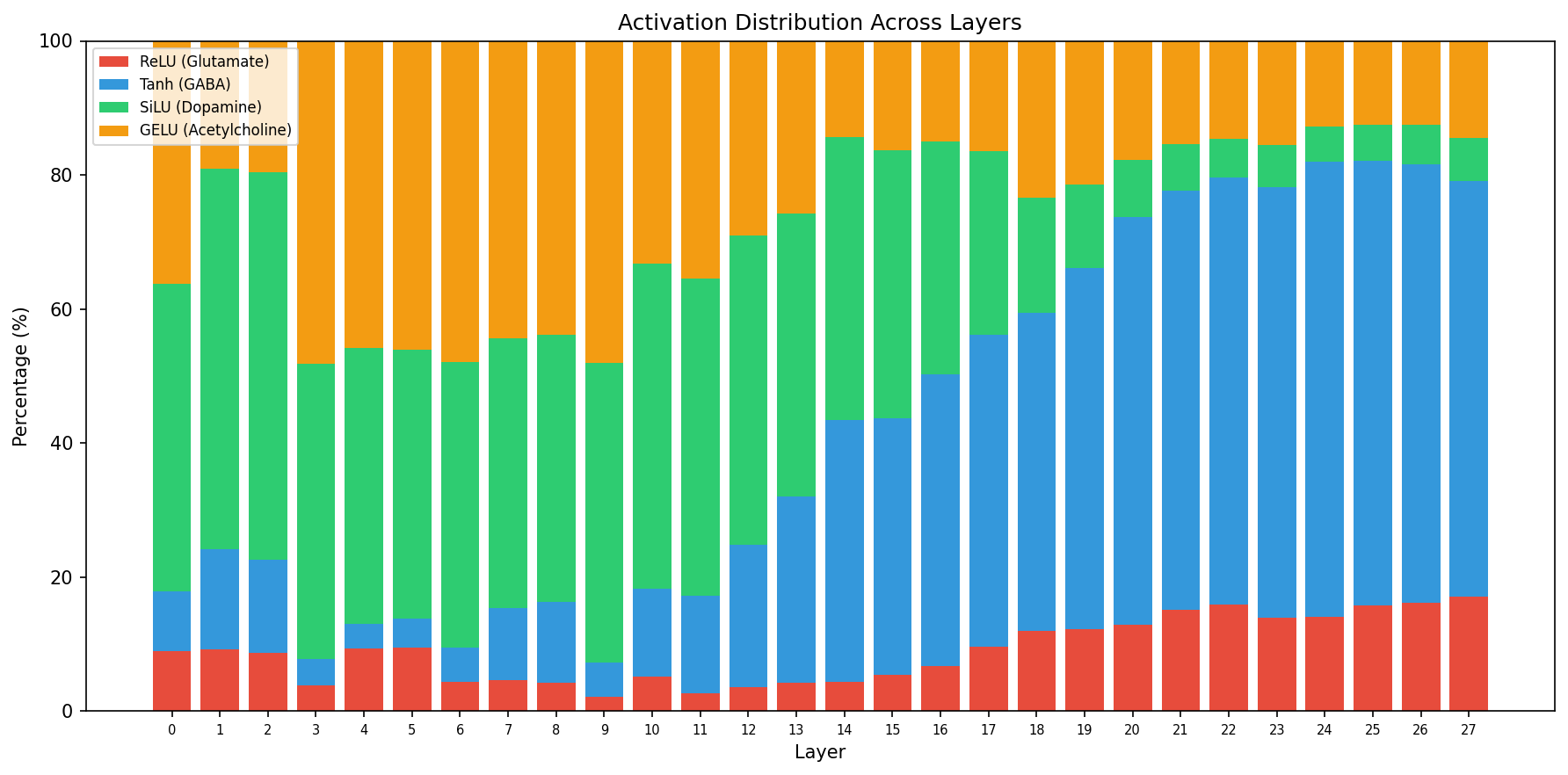}
    \caption{Distribution of preferred activation functions across layers. Early layers prefer GELU; deep layers strongly favor Tanh. All four activations maintain non-trivial representation across the network.}
    \label{fig:layer_dist}
\end{figure}

The layer-wise distribution of activation preferences (Figure~\ref{fig:layer_dist}) quantifies this gradient:

\begin{itemize}
    \item \textbf{Early layers (0--2)}: GELU dominates (${\sim}$35--40\%), with Tanh (${\sim}$15--25\%) and SiLU (${\sim}$15--20\%) as secondary components.
    \item \textbf{Middle layers (3--14)}: Gradual transition. GELU remains the plurality choice, SiLU grows to ${\sim}$15--25\%.
    \item \textbf{Deep layers (15--27)}: Tanh surges dramatically to 50--65\%, becoming the dominant activation. This is striking---the symmetric, bounded activation becomes preferred for deep representational processing.
\end{itemize}

\subsubsection{The Weight Decay Bug and Its Consequences}
\label{sec:weight_decay}

At step ${\sim}$10,000 (${\sim}$51\% through training), we discovered that the static routing preferences $\boldsymbol{\alpha}$ had been inadvertently subjected to weight decay ($\lambda{=}0.1$). Since $\boldsymbol{\alpha}$ is a 2D tensor (shape $[d_{\text{ff}}, K]{=}[4096, 4]$), it was grouped with weight matrices rather than biases. Weight decay continuously penalized deviation from the zero initialization, suppressing static specialization.

This was corrected via an optimizer state transplant at step 10,000---Adam's running statistics (\texttt{exp\_avg}, \texttt{exp\_avg\_sq}) were transferred to the corrected parameter grouping without any loss spike. The critical discovery enabled by this fix was the \emph{dynamic routing entropy} diagnostic: even while $\boldsymbol{\alpha}$ was suppressed, the gate network alone had achieved near-deterministic routing (Section~\ref{sec:static_dynamic}). The static preferences in the neurotransmitter heatmap (Figure~\ref{fig:heatmap}) reflect only the final ${\sim}$9,531 steps of unpenalized learning.

\subsection{Benchmark Performance}

\begin{table}[t]
\centering
\small
\begin{tabular}{@{}llrrrrr@{}}
\toprule
\textbf{Benchmark} & \textbf{Metric} & \textbf{Base} & \textbf{SFT} & \textbf{$\Delta$} & \textbf{Random} & \textbf{Qwen3-0.6B} \\
\midrule
HellaSwag & acc\_norm & 28.51 & 27.84 & $-$0.67 & 25.00 & 41.10 \\
ARC-Easy & acc\_norm & 41.04 & 36.11 & $-$4.93 & 25.00 & 65.60 \\
ARC-Challenge & acc\_norm & 22.27 & 24.15 & $+$1.88 & 25.00 & 33.90 \\
PIQA & acc\_norm & 58.87 & 54.52 & $-$4.35 & 50.00 & 70.00 \\
WinoGrande & acc & 52.17 & 52.72 & $+$0.55 & 50.00 & 58.50 \\
BoolQ & acc & 61.13 & 55.63 & $-$5.50 & 50.00 & 69.70 \\
MMLU-STEM & acc (5-shot) & 25.28 & 28.42 & $+$3.14 & 25.00 & --- \\
LAMBADA & acc & 15.35 & 7.01 & $-$8.34 & ${\sim}$0 & --- \\
OpenBookQA & acc\_norm & 29.00 & 26.80 & $-$2.20 & 25.00 & --- \\
SciQ & acc\_norm & 61.20 & 52.70 & $-$8.50 & 25.00 & --- \\
\midrule
\textbf{Mean} & & \textbf{39.48} & \textbf{36.59} & \textbf{$-$2.89} & & \\
\bottomrule
\end{tabular}
\caption{Benchmark results (\%). The base model scores above random on 8/10 tasks. On the 6 benchmarks with published Qwen3-0.6B-Base scores, PolychromaticLM achieves 62--89\% of Qwen3's performance despite training on 3,600$\times$ fewer tokens.}
\label{tab:benchmarks}
\end{table}

\begin{figure}[h]
    \centering
    \includegraphics[width=0.85\linewidth]{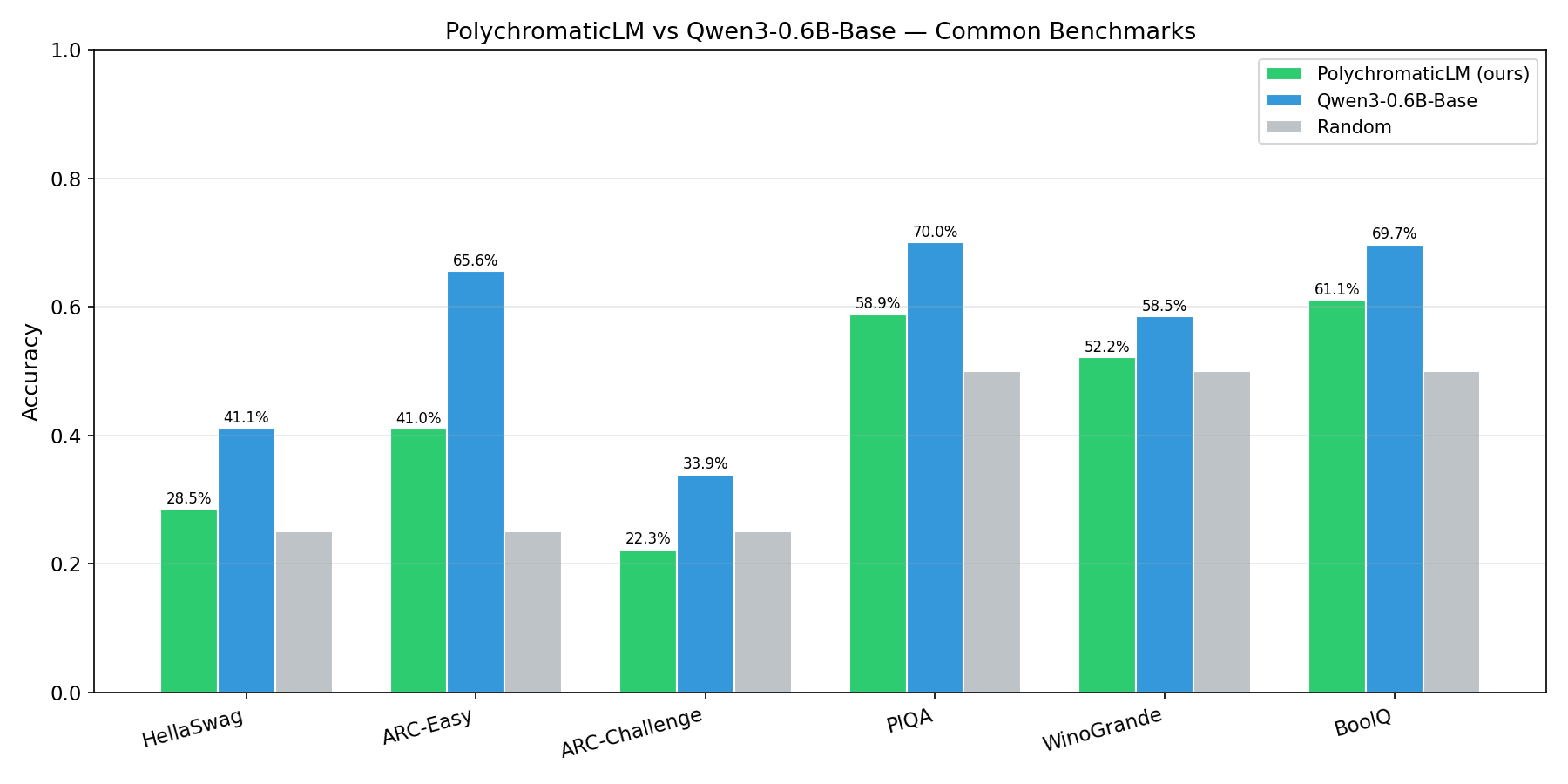}
    \caption{Benchmark comparison between PolychromaticLM base and Qwen3-0.6B-Base. Despite 3,600$\times$ fewer training tokens, PolychromaticLM achieves 62--89\% of Qwen3's scores.}
    \label{fig:benchmark_comparison}
\end{figure}

The base PolychromaticLM achieves meaningful above-random performance on 8/10 benchmarks (Table~\ref{tab:benchmarks}), with particularly strong results on SciQ (61.20\%, $+$36.2pp vs.\ random) and BoolQ (61.13\%, $+$11.1pp). On the six benchmarks where Qwen3-0.6B-Base scores are available, PolychromaticLM achieves 62--89\% of Qwen3's performance (Figure~\ref{fig:benchmark_comparison}). Given that Qwen3 was trained on ${\sim}$36T tokens (3,600$\times$ our budget), this represents strong compute efficiency.

\subsection{Domain Perplexity}

Domain perplexity on held-out data reveals strong specialization aligned with the training mix:

\begin{table}[h]
\centering
\small
\begin{tabular}{@{}lrrr@{}}
\toprule
\textbf{Domain} & \textbf{Training Share} & \textbf{Perplexity} & \textbf{Bits/Token} \\
\midrule
Math & 70\% ($\to$85\%) & 3.56 & 1.83 \\
Code & 5\% & 7.08 & 2.82 \\
STEM & 25\% ($\to$10\%) & 31.93 & 5.00 \\
\bottomrule
\end{tabular}
\caption{Domain perplexity on held-out data (${\sim}$1M tokens per domain). Code achieves lower perplexity than STEM despite 5$\times$ less training data, likely due to lower intrinsic entropy and transfer from mathematical training.}
\label{tab:perplexity}
\end{table}

An unexpected finding is that code perplexity (7.08) is much lower than STEM perplexity (31.93), despite code comprising only 5\% of training versus STEM's 25\%. We attribute this to (1)~the lower intrinsic entropy of Python code compared to diverse scientific prose, and (2)~substantial transfer from mathematical training data (70\%), which shares formal notation and logical structure with code.

\subsection{Fine-Tuning Stability}

\subsubsection{Routing Entropy Preservation}

\begin{figure}[h]
    \centering
    \includegraphics[width=0.85\linewidth]{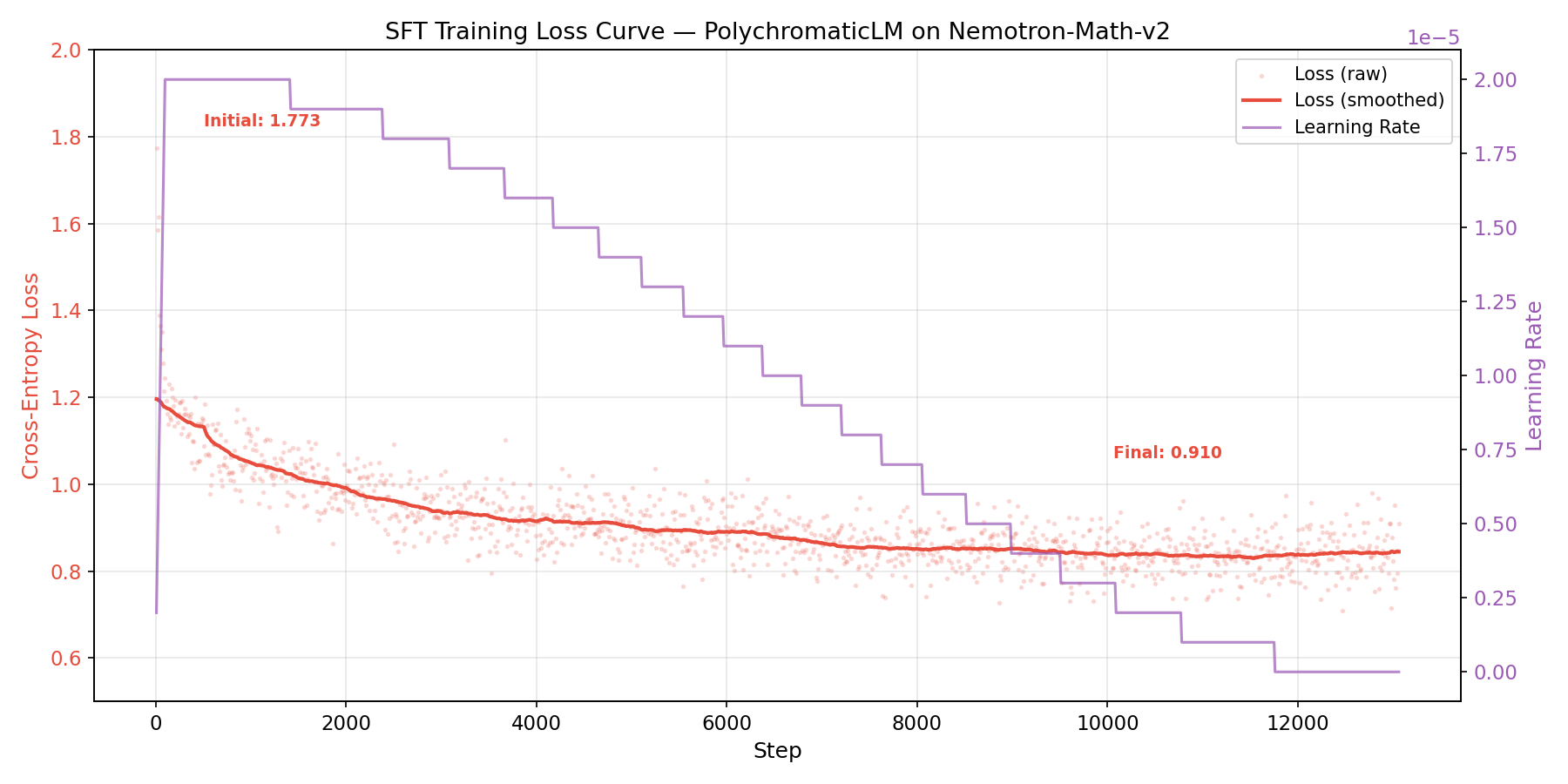}
    \caption{SFT training loss (1.77 $\to$ 0.91, 48.7\% reduction). Routing entropy (not shown) remained constant at 1.386 = $\ln(4)$ throughout.}
    \label{fig:sft_loss}
\end{figure}

The most striking SFT observation is that \textbf{routing entropy remained exactly at $\ln(4) \approx 1.386$ throughout all 13,067 SFT steps} (Figure~\ref{fig:sft_loss}). This means the static routing preferences ($\boldsymbol{\alpha}$) established during pre-training were completely undisturbed by fine-tuning---the routing architecture is a permanent structural feature, not a fragile training artifact.

\subsubsection{Forgetting Analysis}

SFT on mathematical reasoning data produced moderate forgetting across general benchmarks (mean $-$2.89pp, Table~\ref{tab:benchmarks}), with 9/10 tasks remaining above random. The largest regressions occurred on tasks most distant from mathematical reasoning (LAMBADA $-$8.34pp, SciQ $-$8.50pp), while math-adjacent tasks improved (MMLU-STEM $+$3.14pp, ARC-Challenge $+$1.88pp).

\begin{figure}[h]
    \centering
    \includegraphics[width=0.85\linewidth]{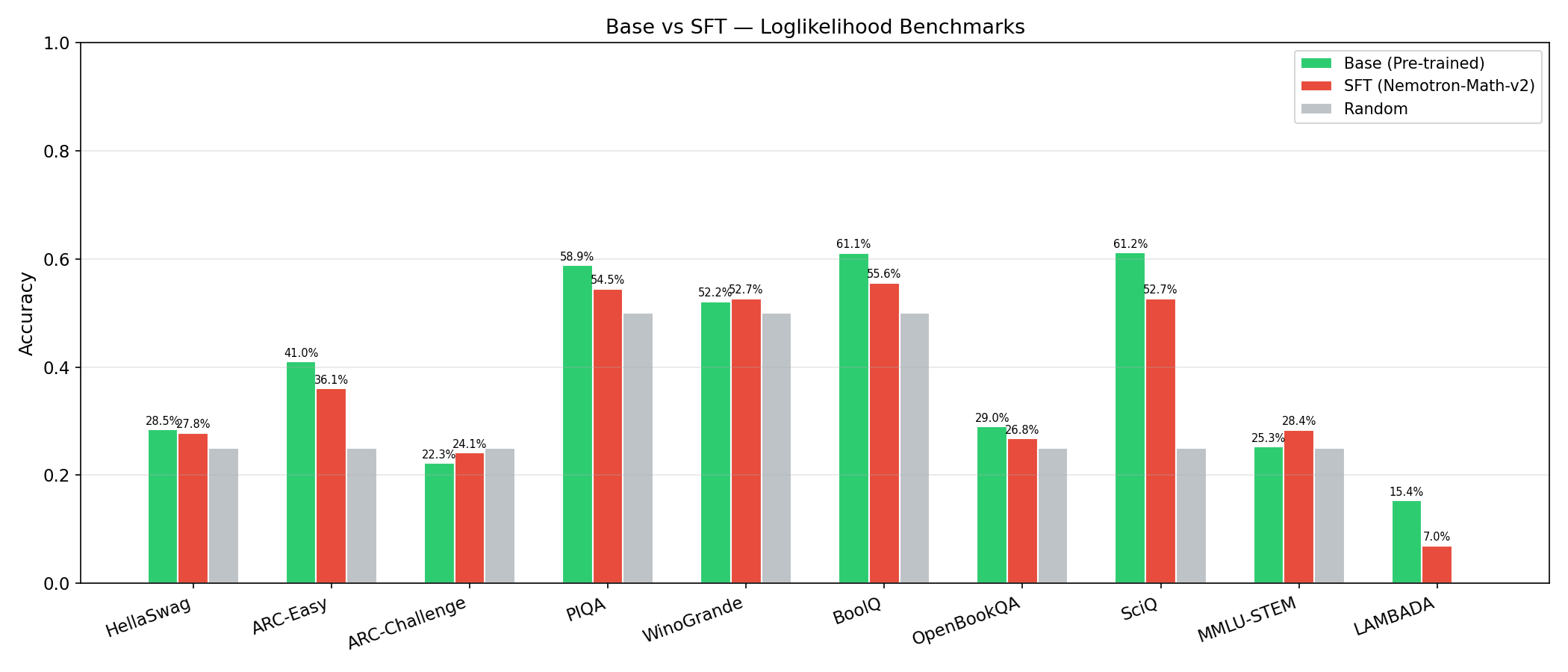}
    \caption{Base vs.\ SFT benchmark comparison. Most benchmarks show moderate regression, with selective improvements on MMLU-STEM and ARC-Challenge.}
    \label{fig:sft_comparison}
\end{figure}

\begin{figure}[h]
    \centering
    \includegraphics[width=0.85\linewidth]{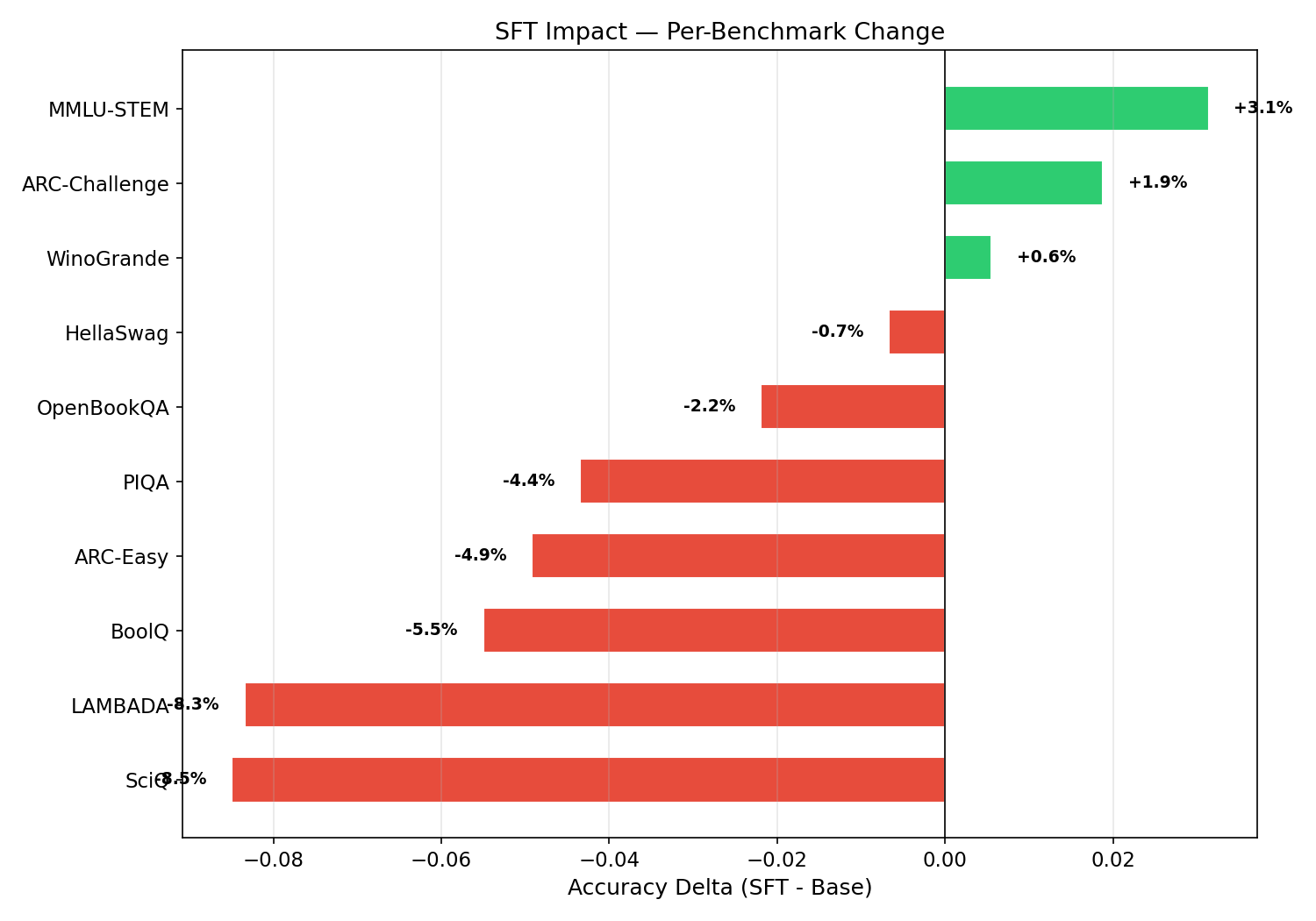}
    \caption{Per-benchmark SFT impact ($\Delta$ from base). MMLU-STEM (+3.14pp) and ARC-Challenge (+1.88pp) improve; LAMBADA ($-$8.34pp) and SciQ ($-$8.50pp) regress most.}
    \label{fig:sft_delta}
\end{figure}

The MMLU-STEM improvement is particularly informative. Detailed subtask analysis shows the largest gains in quantitative subjects: High School Statistics ($+$20.84pp), College Computer Science ($+$18.00pp), College Chemistry ($+$14.00pp), and College Mathematics ($+$11.00pp). This selective transfer---math SFT improving quantitative STEM knowledge---provides indirect evidence that the fine-tuning successfully instilled mathematical reasoning capability.

\section{Analysis and Discussion}

\subsection{Why Does Routing Converge Without Regularization?}

The spontaneous convergence to near-deterministic routing is the most surprising result. In Mixture-of-Experts models, routing collapse (all tokens sent to one expert) is a well-known failure mode that requires explicit load-balancing losses \citep{shazeer2017outrageously, fedus2022switch}. PolyGLU achieves the opposite---diverse, layer-specific specialization---without any auxiliary objective.

We hypothesize that the language modeling loss itself provides sufficient signal for routing specialization. Consider that the gradient of the cross-entropy loss with respect to the routing weights $g_k$ is:
\begin{equation}
    \frac{\partial \mathcal{L}}{\partial g_k} = \frac{\partial \mathcal{L}}{\partial \text{PolyGLU}} \cdot \left[\sigma_k(\mathbf{x} W_{\text{gate}}) \odot (\mathbf{x} W_{\text{up}})\right]
\end{equation}

Each activation function $\sigma_k$ produces a \emph{qualitatively different} gradient signal. Over many training steps, the optimizer discovers that committing to a specific activation for each neuron produces cleaner, more consistent gradients than maintaining a soft mixture. The Gumbel-Softmax temperature annealing amplifies this effect: as $\tau$ decreases, the routing distribution sharpens, further reinforcing committed selections via a positive feedback loop.

\subsection{The Static-Dynamic Separation}
\label{sec:static_dynamic}

A key architectural insight emerged from the weight decay bug (Section~\ref{sec:weight_decay}): the dynamic routing pathway alone is sufficient for near-deterministic routing. Even while $\boldsymbol{\alpha}$ was suppressed by weight decay for 10,000 steps, the gate network achieved mean dynamic entropy of 0.58\% of maximum. This demonstrates that:

\begin{enumerate}
    \item The gate network---a simple 2-layer MLP processing mean-pooled hidden states---has sufficient expressive power to make confident, per-neuron routing decisions.
    \item The routing signal is primarily \emph{contextual}: the network learns which activation to use based on what it is currently processing, not solely from fixed per-neuron preferences.
    \item The static component $\boldsymbol{\alpha}$ is architecturally present but functionally subordinate. Its primary role may be to provide a warm-starting bias that accelerates routing convergence.
\end{enumerate}

\subsection{Implications for Activation Function Design}

The emergent depth-dependent specialization (GELU-early, Tanh-deep) challenges the implicit assumption that a single activation is optimal across all layers. This finding suggests several directions:

\begin{itemize}
    \item \textbf{Heterogeneous fixed activations}: One could train a PolyGLU model, observe the converged routing pattern, and then ``distill'' it into a standard transformer with different fixed activations per layer. This would capture the benefits of adaptive routing at zero inference cost.
    \item \textbf{Activation search}: PolyGLU routing patterns could serve as a data-driven method for activation function selection, analogous to how neural architecture search discovers optimal layer configurations.
    \item \textbf{Deeper investigation}: The preference for Tanh in deep layers is particularly intriguing. Tanh's bounded, symmetric output may provide implicit regularization in deep representations, preventing activation magnitudes from growing unboundedly. This deserves dedicated investigation.
\end{itemize}

\subsection{Limitations and Future Work}

\paragraph{No SwiGLU baseline.} The most important limitation is the absence of a controlled ablation against a vanilla SwiGLU model with identical architecture, data, and training budget. Without this comparison, we cannot definitively attribute the observed behaviors to PolyGLU versus the specific training setup. Budget constraints prevented training a second 597M-parameter model; this is the highest-priority follow-up experiment.

\paragraph{No GSM8K evaluation.} Generation-based evaluation without KV cache proved prohibitively expensive (${\sim}$9 hours for 1,319 examples on A100). Approximately 98\% of examples were processed before the evaluation was terminated, but the lm-evaluation-harness writes results only upon completion. Optimized inference code with KV cache and complete GSM8K evaluation results will be released separately in a follow-up publication.

\paragraph{Scale.} All experiments were conducted at ${\sim}$600M parameters and ${\sim}$10B tokens. Scaling behavior---whether the routing patterns persist, intensify, or change qualitatively at larger scales---remains an open question.

\paragraph{Activation palette.} We used $K{=}4$ activations chosen heuristically. Systematic exploration of the palette size and composition (e.g., including Mish, squared ReLU, or learned parametric activations) could yield further improvements.

\paragraph{Inference efficiency.} At inference time, PolyGLU requires computing all $K$ activations before selecting one, adding computational overhead proportional to $K$. For production deployment, the converged routing pattern could be ``frozen'' into a static mapping, eliminating the runtime cost entirely.

\section{Conclusion}

We have presented PolyGLU, a simple modification to the transformer feed-forward block that replaces fixed activation functions with state-conditional activation routing. Training a 597M-parameter model on ${\sim}$10B tokens reveals striking emergent behavior: the routing mechanism spontaneously converges to near-deterministic selections (0.030\% of maximum entropy) without any explicit regularization, and a clear depth-dependent specialization pattern emerges---early layers prefer GELU while deep layers strongly favor Tanh.

The routing architecture proves remarkably robust: it survives fine-tuning with zero entropy drift, adds only 0.23\% parameter overhead, and produces interpretable neurotransmitter maps that reveal how the network organizes its computational resources. Three layers (9, 16, 17) maintain elevated routing entropy, suggesting the model has discovered computational flexibility points where activation diversity is beneficial.

These findings demonstrate that the ``one activation fits all'' assumption in modern transformers is suboptimal---different network depths benefit from different nonlinear transformations, and gradient-based optimization can discover these specialization patterns given the freedom to do so. We hope that PolyGLU and the PolychromaticLM model inspire further investigation into adaptive, biologically-inspired computation in neural architectures.

\paragraph{Reproducibility.} All code, model weights, training logs, and evaluation scripts are available at \url{https://github.com/danielxmed/PolyGLU}. The entire project was conducted on a single A100 80GB GPU at a total cost of ${\sim}$\$346.

\bibliography{references}

\clearpage
\appendix

\section{Additional Training Figures}

\begin{figure}[H]
    \centering
    \includegraphics[width=0.85\linewidth]{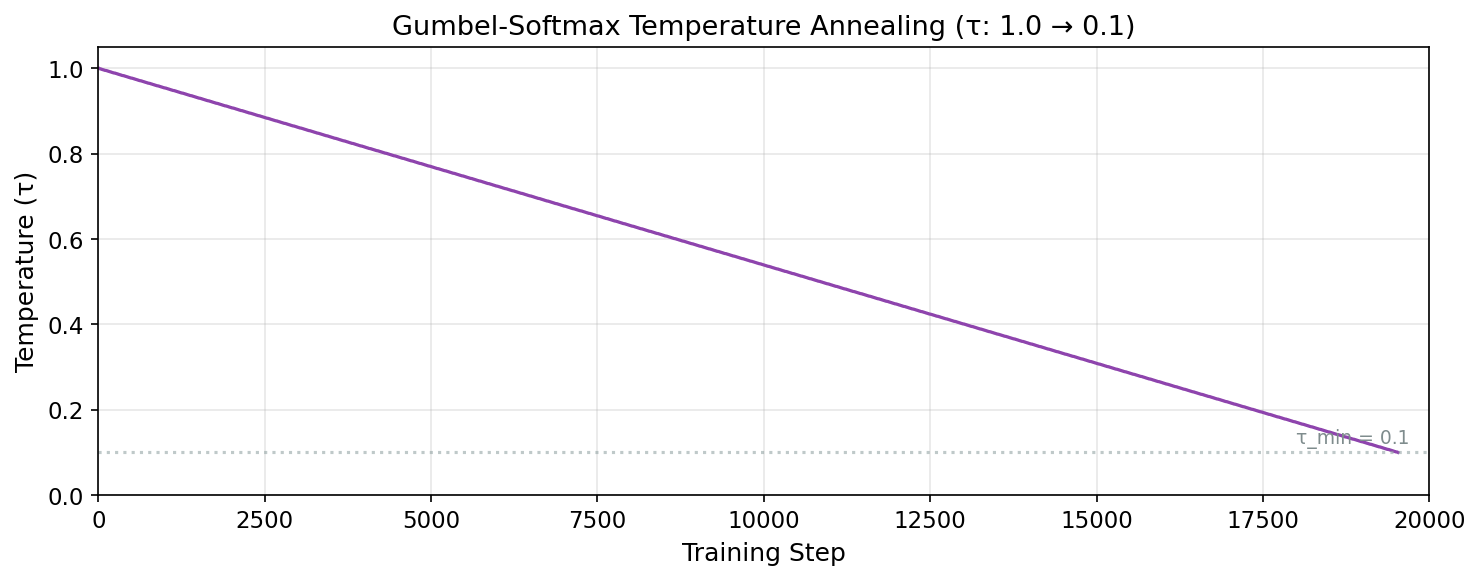}
    \caption{Gumbel-Softmax temperature annealing schedule. $\tau$ decreases linearly from 1.0 to 0.1 over the full training run, transitioning from exploratory to committed routing.}
    \label{fig:tau}
\end{figure}

\begin{figure}[H]
    \centering
    \includegraphics[width=0.85\linewidth]{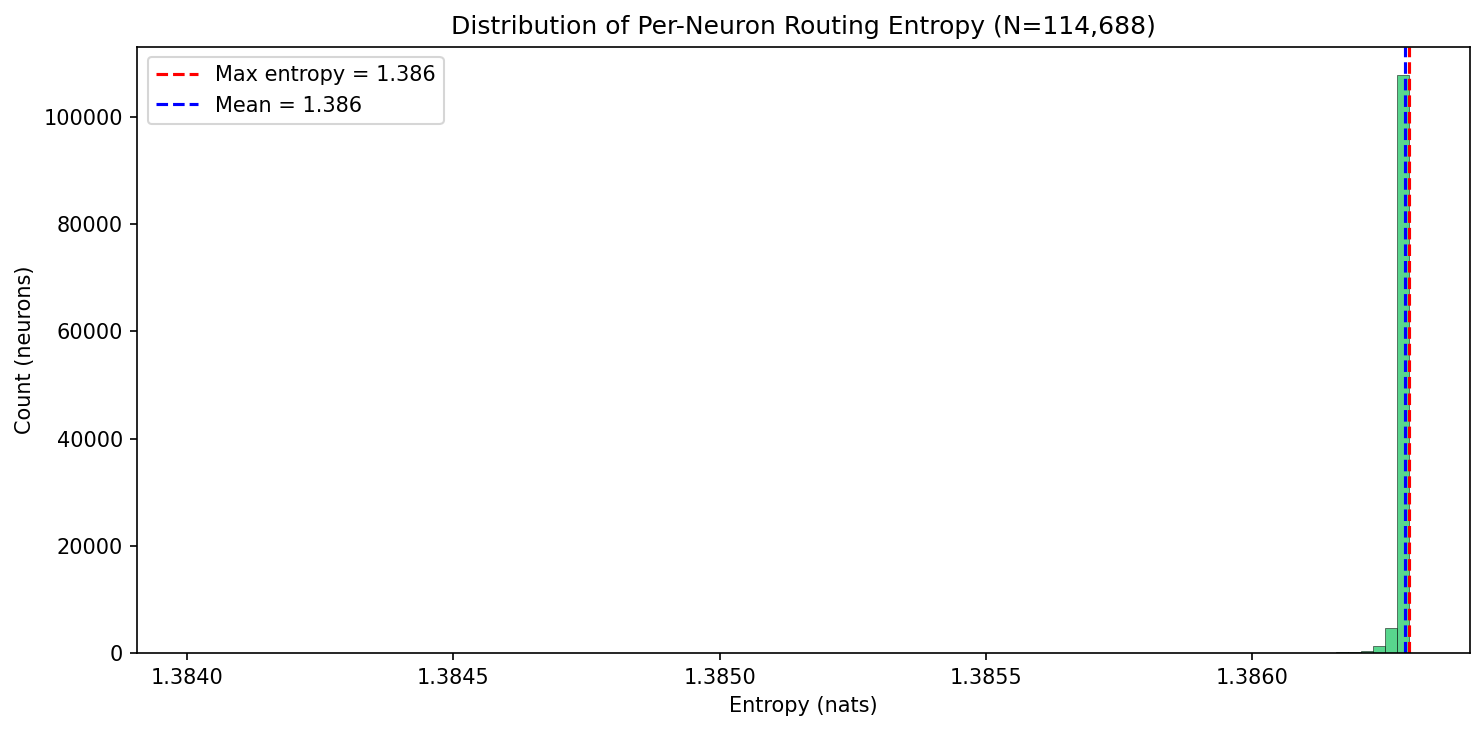}
    \caption{Distribution of per-neuron \emph{static} routing entropy ($H(\text{softmax}(\boldsymbol{\alpha}))$) at convergence. All neurons cluster at the maximum $\ln(4) \approx 1.386$, reflecting the suppression of $\boldsymbol{\alpha}$ by weight decay during the first half of training (Section~\ref{sec:weight_decay}). Actual routing decisions are near-deterministic via the dynamic pathway (Figure~\ref{fig:entropy_final}).}
    \label{fig:entropy_hist}
\end{figure}

\begin{figure}[H]
    \centering
    \includegraphics[width=0.85\linewidth]{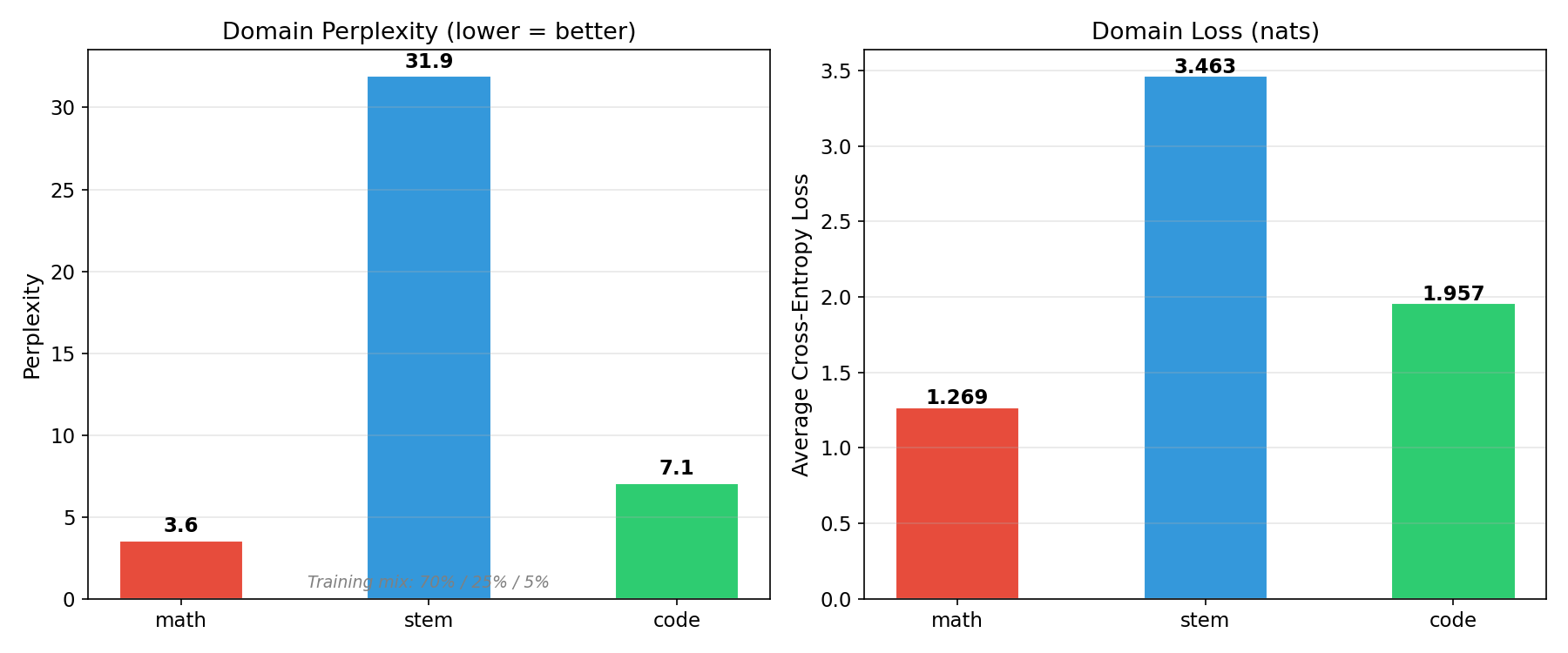}
    \caption{Domain perplexity comparison across Math, Code, and STEM held-out data. Code achieves lower perplexity than STEM despite 5$\times$ less training data.}
    \label{fig:domain_perplexity}
\end{figure}

\section{Detailed MMLU-STEM Subtask Scores}

\begin{table}[H]
\centering
\small
\begin{tabular}{@{}lrrr@{}}
\toprule
\textbf{Subtask} & \textbf{Base (\%)} & \textbf{SFT (\%)} & \textbf{$\Delta$ (pp)} \\
\midrule
Abstract Algebra & 30.00 & 22.00 & $-$8.00 \\
Anatomy & 34.07 & 25.19 & $-$8.88 \\
Astronomy & 25.00 & 28.29 & $+$3.29 \\
College Biology & 22.22 & 26.39 & $+$4.17 \\
College Chemistry & 19.00 & 33.00 & $+$14.00 \\
College Computer Science & 15.00 & 33.00 & $+$18.00 \\
College Mathematics & 25.00 & 36.00 & $+$11.00 \\
College Physics & 19.61 & 28.43 & $+$8.82 \\
Computer Security & 24.00 & 22.00 & $-$2.00 \\
Conceptual Physics & 24.26 & 21.70 & $-$2.56 \\
Electrical Engineering & 31.72 & 26.21 & $-$5.51 \\
Elementary Mathematics & 26.72 & 26.98 & $+$0.26 \\
High School Biology & 26.13 & 33.23 & $+$7.10 \\
High School Chemistry & 25.12 & 25.62 & $+$0.50 \\
High School Computer Science & 33.00 & 30.00 & $-$3.00 \\
High School Mathematics & 25.93 & 24.44 & $-$1.49 \\
High School Physics & 25.17 & 34.44 & $+$9.27 \\
High School Statistics & 20.83 & 41.67 & $+$20.84 \\
Machine Learning & 23.21 & 19.64 & $-$3.57 \\
\midrule
\textbf{Aggregate} & \textbf{25.28} & \textbf{28.42} & \textbf{$+$3.14} \\
\bottomrule
\end{tabular}
\caption{Detailed MMLU-STEM subtask comparison. Math SFT selectively improves quantitative subjects (High School Statistics $+$20.84pp, College Mathematics $+$11.00pp) while non-quantitative subjects show mixed results.}
\label{tab:mmlu_detail}
\end{table}

\section{Key Training Milestones}

\begin{table}[H]
\centering
\small
\begin{tabular}{@{}rrrll@{}}
\toprule
\textbf{Step} & \textbf{Tokens} & \textbf{Loss} & \textbf{$\tau$} & \textbf{Phase} \\
\midrule
10 & 5.2M & 12.13 & 1.00 & Initial \\
100 & 52.4M & 9.23 & 1.00 & Early learning \\
500 & 262M & 5.46 & 0.98 & Rapid descent \\
1,000 & 524M & 4.17 & 0.95 & Warmup \\
2,000 & 1.05B & 3.50 & 0.91 & Warmup end \\
5,000 & 2.62B & 2.25 & 0.77 & Steady optimization \\
10,000 & 5.24B & 2.26 & 0.54 & Mid-training fix \\
15,000 & 7.86B & 1.68 & 0.31 & Annealing onset \\
19,531 & 10.24B & \textbf{1.31} & 0.10 & Final \\
\bottomrule
\end{tabular}
\caption{Key pre-training milestones. The mid-training intervention at step 10,000 (weight decay bug fix) produced no loss discontinuity.}
\label{tab:milestones}
\end{table}

\end{document}